\documentclass[letterpaper, 10 pt, conference]{ieeeconf}  

\IEEEoverridecommandlockouts                              

\usepackage{amsmath} 
\usepackage{amssymb}  
\usepackage{booktabs}
\usepackage{tabularx}
\usepackage{multirow}
\usepackage{multicol}
\usepackage{graphicx}
\usepackage{subcaption}
\usepackage{float}
\usepackage{pifont}

\title{\LARGE \bf
CLASH: Collision Learning via Augmented Sim-to-real Hybridization to Bridge the Reality Gap
}

\author{ Haotian He$^{1}$, Ning Guo$^{2}$, Siqi Shi$^{3}$, Qipeng Liu$^{2}$, Wenzhao Lian$^{2,\dagger}$
\thanks{$^{1}$School of Mathematical Sciences, Peking University. $^{2}$School of Artificial Intelligence, Shanghai Jiao Tong University. $^{3}$School of Mathematical Sciences, Beijing Normal University.$^{\dagger}$Corresponding Author.}
}

\begin{document}

\maketitle
\thispagestyle{empty}
\pagestyle{empty}

\begin{abstract}

The sim-to-real gap, particularly in the inaccurate modeling of contact-rich dynamics like collisions, remains a primary obstacle to deploying robot policies trained in simulation. Conventional physics engines often trade accuracy for computational speed, leading to discrepancies that prevent direct policy transfer. To address this, we introduce Collision Learning via Augmented Sim-to-real Hybridization (CLASH),  a data-efficient framework that learns a parameter-conditioned impulsive collision surrogate model and integrates it as a plug-in module within a standard simulator. CLASH first distills a base model from an imperfect simulator (MuJoCo) using large-scale simulated collisions to capture reusable physical priors. Given only a handful of real collisions (e.g., 10 samples), it then (i) performs gradient-based identification of key contact parameters and (ii) applies small-step, early-stopped fine-tuning to correct residual sim-to-real mismatches while avoiding overfitting. The resulting hybrid simulator not only achieves higher post-impact prediction accuracy but also reduces the wall-clock time of collision-heavy CMA-ES search by 42-48\% compared to MuJoCo. We demonstrate that policies obtained with our hybrid simulator transfer more robustly to the real world, doubling the success rate in sequential pushing tasks with reinforcement learning and significantly increase the task performance with model-based control.

\end{abstract}

\section{INTRODUCTION}

In recent years, simulation environments have seen explosive growth in capability, becoming a cornerstone of robotic strategy development. Unlike empirical data collection in the real world, which is often costly, time-consuming, and poses safety risks, training in simulation allows for rapid and scalable generation of diverse experiences in a controlled, risk-free setting\cite{zhao2020sim}. This efficiency and safety make simulators indispensable for pretraining complex policies before deployment on physical systems. Modern engines such as MuJoCo\cite{todorov2012mujoco} can capture rich rigid-body and contact dynamics, while GPU-parallel platforms like NVIDIA Isaac Lab\cite{isaaclab_docs} make it feasible to train at massive scale with thousands of synchronized environments. These simulators provide a low-cost way to train a preliminary policy with state-of-art reinforcement learning algorithms like PPO\cite{schulman2017proximal} or SAC\cite{haarnoja2018soft}. However, the “sim-to-real” gap, stemming from unmodeled effects, sensing latencies, inaccurate identified parameters and other mismatches, often prevents policies learned in simulation from transferring directly to physical robots.

Among various physical processes, collisions are particularly challenging to simulate accurately due to their discontinuous nature and sensitivity to frictional effects. Formally, rigid-body collisions with friction are governed by non-linear complementarity problems (NCPs)\cite{howell2022dojo}, which are hard to solve both accurately and efficiently. To maintain real-time performance, many simulators\cite{coumans2016pybullet,nvidia_physics_sdk,tedrake2019drake, werling2021fast} adopt various relaxations and approximations, such as penalty-based contacts or linearized friction cones. While computationally efficient, these relaxations often introduce errors in angular velocity change, frictional dissipation, and post-impact trajectories. Since many manipulation tasks such as pushing or striking rely critically on collision dynamics, reducing this sim-to-real gap in collision modeling is essential.

To mitigate sim-to-real gap mentioned above, many methods are proposed such as improving system identification\cite{memmel2024asid,chen2024learning,shi2025real,wang2025embodiedreamer}, domain randomization\cite{tobin2017domain}, domain adaptation\cite{bousmalis2018using} or augmenting simulators with learned models\cite{heiden2021neuralsim,gao2024sim,ajay2018augmenting,fazeli2017learning}. However, directly learning an augmented model often requires a large amount of real-world data. To improve data efficiency, a neural surrogate model can first be pre-trained on simulation data to approximate the simulator’s dynamics. This can be viewed as distillation from a physics engine: rather than learning an unconstrained black-box correction, the simulator’s implicit physical priors are distilled into a differentiable surrogate model that is explicitly conditioned on a small set of physically interpretable and identifiable parameters. This design enables gradient-based system identification from only a handful of real interactions by optimizing these parameters with the network weights fixed, followed by small-step, early-stopped fine-tuning to capture the remaining non-parametric mismatch. In this way, the sim-to-real discrepancy is decomposed into an interpretable parametric component and a limited residual, improving both data efficiency and interpretability compared to purely residual-learning approaches.

\begin{figure}
    \centering
    \includegraphics[width=\linewidth]{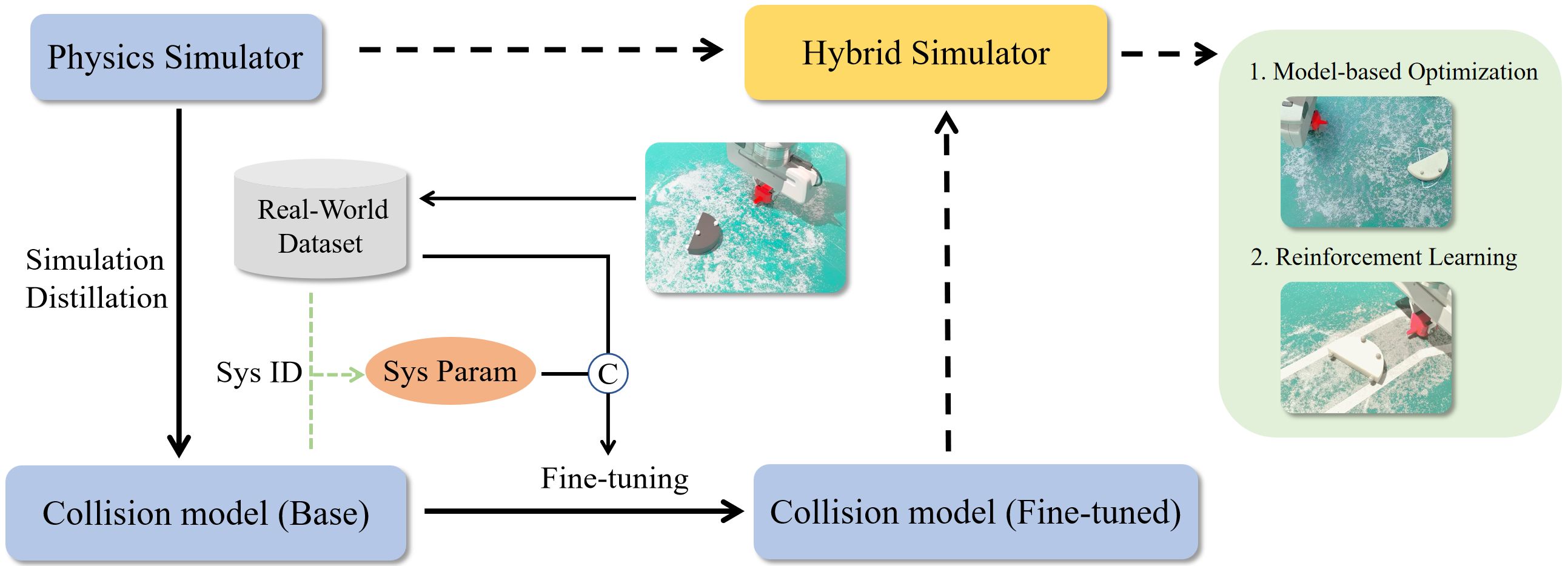}
    \caption{Overview of our CLASH framework and learning downstream tasks with the hybrid simulator.}
    \label{fig_overview}
\end{figure}

\begin{table*}
    \centering
    \caption{Comparison with Prior Sim-to-Real Modeling Methods}
    \label{table_0}
    \resizebox{\linewidth}{!}{
    \begin{tabular}{cccccc}
    \toprule
    Method & Simulator Dependency & System Identification & Hybrid Model & Real Data Requirement & Accuracy\\
    \midrule
    MuJoCo\cite{todorov2012mujoco}& \ding{51} &\ding{51}(0th-order)&\ding{55}&Few& Low\\
    MJX\cite{mjx_docs}& \ding{51}& \ding{51}(1st-order)&\ding{55}&Few& Low\\
    Residual Learning\cite{gao2024sim}& \ding{51}&\ding{55}&\ding{51}&Few&High\\
    NeuralSim\cite{heiden2021neuralsim}& \ding{55}&\ding{55}&\ding{55}&Many&High\\
    CLASH(ours)& \ding{51}&\ding{51}(1st-order)&\ding{51}&Few& High\\
    \bottomrule
    \end{tabular}}
\end{table*}
Building on this insight, we introduce a hybrid simulation framework and choose collision dynamics as the representative use case to demonstrate the learning procedure. Specifically, we propose our framework, Collision Learning via Augmented Sim-to-real Hybridization (CLASH), as shown in Fig. \ref{fig_overview}. The hybrid simulator is not only more accurate than the traditional simulator MuJoCo, but also significantly reduces the computation time associated with collision calculation.

Our contribution can be summarized as follows:
\begin{itemize}
    \item We introduce CLASH, a data-efficient hybridization framework that distills simulator priors into a differentiable surrogate model conditioned on identifiable contact parameters, enabling gradient-based system identification from a handful of real interactions and small-step adaptation to correct residual sim-to-real mismatch.    
    \item We build a plug-in hybrid simulator by integrating the learned surrogate model as a collision module within MuJoCo, improving post-impact prediction fidelity in our target collision settings while reducing the wall-clock cost of collision-heavy simulation and planning.
    \item We demonstrate that the resulting hybrid simulator consistently improves downstream performance in both model-based control and reinforcement learning, leading to more reliable sim-to-real transfer and higher real-world task success.
\end{itemize}

\section{RELATED WORK}

\subsection{Physics Simulator}
In recent years, various physics simulators have provided high-performance rigid-body dynamics and contact modeling for robotics and reinforcement learning~\cite{todorov2012mujoco,isaaclab_docs,authors2024genesis,coumans2016pybullet,geng2025roboverse}. 
MuJoCo~\cite{todorov2012mujoco} offers efficient contact dynamics and actuator models and is widely used as a baseline for model-based control and RL.
Isaac Lab~\cite{isaaclab_docs} supports GPU-parallel simulation for large-scale policy training and evaluation.
PyBullet~\cite{coumans2016pybullet} provides a lightweight and accessible simulator that is commonly used for rapid prototyping.
Genesis~\cite{authors2024genesis} aims to provide a unified physics engine with broad robotics support.
RoboVerse~\cite{geng2025roboverse} integrates multiple simulators into a unified framework to facilitate scalable learning and sim-to-real transfer.
While these platforms excel in speed, stability, and community adoption, most do not natively support automatic differentiation, limiting their use for gradient-based system identification and data-efficient optimization.

To utilize gradient information, differentiable simulators and toolkits such as TDS~\cite{heiden2021neuralsim}, MJX~\cite{mjx_docs}, Warp~\cite{warp2022}, and GradSim~\cite{jatavallabhula2021gradsim} provide end-to-end differentiation through dynamics (and sometimes contact and rendering).
MJX~\cite{mjx_docs} reimplements MuJoCo in JAX, enabling differentiable simulation with GPU acceleration.
Warp~\cite{warp2022} offers a high-performance GPU framework for differentiable simulation and graphics, supporting flexible custom physics.
GradSim~\cite{jatavallabhula2021gradsim} combines differentiable multi-physics with differentiable rendering for system identification and visuomotor control.
TDS/NeuralSim-style approaches~\cite{heiden2021neuralsim} augment differentiable simulators with learned components to improve modeling capacity.
Despite enabling gradient-based optimization, these differentiable pipelines can face trade-offs in efficiency, numerical stability, or modeling fidelity in complex contact-rich interactions.

\subsection{Contact Modeling}

Contact dynamics in rigid-body simulation is often formulated as a nonlinear complementarity problem (NCP)\cite{howell2022dojo}. However, solving NCPs can be computationally expensive and prone to numerical instability, especially in complex contact scenarios. To achieve efficient and stable simulation, modern physics engines typically introduce various relaxations to the original problem. A typical approach is to approximate the nonlinear second-order friction cone with a linearized cone, thereby reducing the problem to a linear complementarity problem (LCP)\cite{tedrake2019drake, coumans2016pybullet, werling2021fast}. However, this relaxation may cause a misalignment between the contact friction force and the true velocity direction. Another widely used strategy is to soften the complementarity constraints, effectively replacing hard contacts with penalty-based compliant contacts\cite{todorov2012mujoco, tedrake2019drake}. While computationally efficient, this approach inevitably introduces non-physical penetration. These relaxations reflect a fundamental trade-off between accuracy and efficiency. Despite the resulting sim-to-real discrepancies, modern simulators generally provide a useful approximation of post-impact states, which can subsequently be corrected with real-world data.

\subsection{Closing the Sim2real Gap}

A common strategy to close the sim-to-real gap is Domain Randomization\cite{tobin2017domain}, which exposes policies to a wide range of simulated environmental and physical variations during training. But the method is limited if the policy strongly relies on the environmental parameters. Another useful method is domain adaptation\cite{bousmalis2018using}, which aligning features or visual representations in simulation to real world to mitigate the gap. 

Accurate parameter identification is another common approach to build a simulation environment that is closer to reality. \cite{chen2024learning} uses Warp\cite{warp2022}, a differentiable simulator and robot proprioception to deduce object parameters. \cite{memmel2024asid} trains an exploration policy to interact with the object and estimates the parameters while \cite{shi2025real} builds a Real-to-Sim-to-Real loop framework and updates estimated parameters iteratively. The most relative one to our work is \cite{wang2025embodiedreamer}, which trains a surrogate model with simulators to identify parameters. However, they still infer the future states with physics simulators rather than the surrogate model, thus fail to eliminate complex, unmodeled effects that cannot be captured by tuning a fixed set of physical parameters.

Beyond system identification, augmenting simulators with neural networks has emerged as a prominent direction to compensate for unmodeled effects. For example, \cite{gao2024sim} learns the residual dynamics between simulation and the real world to improve policy transfer, though the model bias may hinder efficient learning. Another line of work directly leverages data-driven models: \cite{golemo2018sim} employs an LSTM to correct policies trained in simulation, while \cite{ajay2018augmenting} further models the sim-to-real gap and its uncertainty using variational recurrent neural networks. \cite{heiden2021neuralsim} incorporates a differentiable simulator to identify system parameters and neural networks are trained to model the residual effects. However, unstable gradients in contact-rich scenarios limit the broader applicability of this framework. To extend simulator augmentation with learned components under a limited real-data budget, we propose CLASH, a data-efficient hybridization framework. Table~\ref{table_0} summarizes the key differences between CLASH and prior sim-to-real modeling approaches.

\section{METHOD}

\subsection{System Overview}
\begin{figure*}
    \centering
    \includegraphics[width=0.9\linewidth]{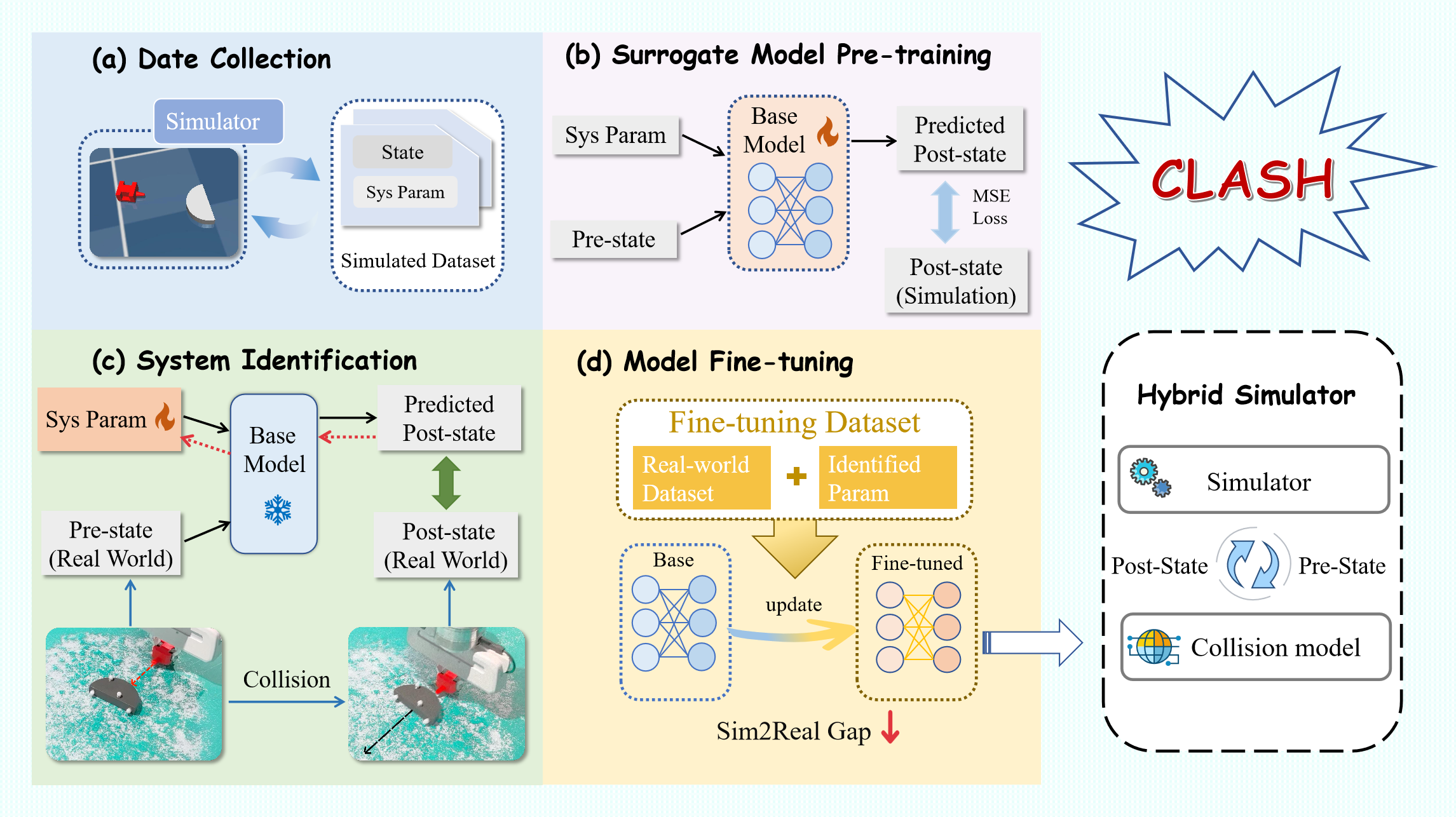}
    \caption{Our framework Collision Learning via Augmented Sim-to-real Hybridization (CLASH): First collect collision dataset from physics simulator and pre-train a base model to learn the collision dynamics, then adopt the model to identify the system parameters for real-world collision. Subsequently, fine-tune with real-world dataset to obtain a collision model. Finally, integrate the collision model with the physics simulator to build a hybrid simulator that is close to reality. }
    \label{fig_CLASH}
\end{figure*}
In this section, we present an overview of our framework, Collision Learning via Augmented Sim-to-real Hybridization (CLASH), as illustrated in Fig. \ref{fig_CLASH}. Since impulsive collision outcomes are among the least accurately modeled components in common physics engines, we augment the simulator by learning a collision surrogate model and integrating it as a plug-in module.
We define collisions as short-duration impulsive contacts between objects, in contrast to continuous sliding interactions on a tabletop. When collisions take place with ground contact maintained, we model post-impact velocities as the combined outcome of object–object impulse and ground reaction.

CLASH follows a three-stage procedure to achieve both data efficiency and high fidelity. (i) Simulation distillation: we pre-train a differentiable surrogate model on a large set of simulated collisions to distill reusable priors from the physics engine. (ii) System identification and adaptation: given only a small set of real impacts, we first identify a compact set of collision-related physical parameters by optimizing them with the network weights fixed, and then apply small-step, early-stopped fine-tuning to capture the remaining sim-to-real mismatch. (iii) Hybrid simulation: we integrate the adapted surrogate model into the simulator and invoke it only when a collision is detected, while retaining the original engine for the remaining dynamics. The resulting hybrid simulator provides more accurate post-impact predictions in real-world settings and can be used for downstream collision-driven manipulation, including reinforcement learning and derivative-free model-based search.



\subsection{Training Surrogate Base Model}

We denote the system parameters as $P=(\boldsymbol{\mu},\boldsymbol{e})$, where $\boldsymbol{\mu}$ is the vector of friction coefficients and $\boldsymbol{e}$ represents the damping ratio from MuJoCo’s solref parameter, which controls the restitution behavior in collisions. Both vectors include values for object–object contacts as well as for object–ground contacts. 
Given two geometry meshes to collide, we randomly set experimental parameters $P$, collision velocity and impact point. Then we define the pre-collision state $S_{\text{pre}}$ according to the chosen velocity and impact point. The physics simulator then executes the collision process and outputs the corresponding post-collision state $S_{\text{post}}$:  
\begin{equation}
    S_{\text{post}} = \text{Sim}(S_{\text{pre}},P).
\end{equation}

By repeating this procedure, we collect a dataset of collision pairs $\tau:=(S_{\text{pre}}, P,S_{\text{post}})$, which constitutes the simulation dataset $D_{\text{sim}}$. A surrogate base model $f_{\theta}(S_{pre}, P)$ is then trained to capture the collision dynamics for the given meshes. The model takes the pre-collision state and system parameters as input and predicts the post-collision state. Since our focus is on the mapping from velocities before and after the collision rather than the full transient process, we parameterize $f_{\theta}$ as a three-layer multilayer perceptron (MLP). The model is optimized by minimizing the mean squared error (MSE) between the predicted and simulated post-collision states:
\begin{equation}
    \mathcal{L}_{\text{base}} = \frac{1}{|D_{\text{sim}}|}\sum_{ \tau\in D_{\text{sim}}}||f_{\theta}(S_{\text{pre}}, P)-S_{\text{post}}||^2.
\end{equation}

This surrogate model approximates the non-linear collision dynamics with a differentiable neural network. While it can accurately predict post-collision states within the simulator, it inevitably inherits the simulator’s modeling and numerical errors. Consequently, a small amount of real-world data is required to adapt the surrogate and obtain reliable predictions for real-world collisions. Although the surrogate model is instantiated per geometry pair, training is designed to remain lightweight: offline simulator data can be generated efficiently by broad randomization over pre-collision states and collision-related parameters and the resulting differentiable model enables efficient system identification and adaptation as described next.

\subsection{System Identification And Fine-tuning}

To adapt the base model to real-world scenarios, we first collect collision data $D_{\text{real}}=\{(S_{\text{pre}},S_{\text{post}})\}$ with a motion capture system. For brevity, denote a single real-world sample as $\hat{\tau} := (S_{\text{pre}}, S_{\text{post}})$. Before fine-tuning the base model, we need to identify the system parameters. When collecting data from a real scene, we assume that the parameters remain constant to maintain consistency for system identification. Since the motion capture data include complete sliding trajectories, for simplicity, the friction coefficients between objects and the ground are efficiently determined using linear regression. For the remaining parameters, we leverage the differentiable simulation-distilled base model.
With frozen weights in the model, we optimize the system parameters to minimize the MSE loss:
\begin{equation}
    \mathcal{L}_{\text{sys}}=  \frac{1}{|D_{\text{real}}|}\sum_{\hat{\tau}\in D_{\text{real}}}|| f_{\theta}(S_{\text{pre}}, P) - S_{\text{post}}||^2.
\end{equation}
    
Here $f_{\theta}$ is a MLP, so we can apply gradient-based optimization to efficiently obtain the optimal parameters.

The identified parameters $P_{\text{real}}:= \arg\min_P\  \mathcal{L}_{\text{sys}}$ can be directly used in the physics simulator, as done in \cite{wang2025embodiedreamer}. But to further leverage real-world collision data and reduce the sim-to-real gap, we additionally fine-tune the surrogate base model using $D_{\text{real}}$ with $P_{\text{real}}$ fixed. To avoid overfitting, we perform small-step and early-stopped gradient updates (e.g., 10 iterations). Conceptually, this adaptation learns data-driven residuals that capture unmodeled real-world effects (e.g., subtle frictional dissipation and compliance). The fine-tuned model remains faithful to the original simulator’s structure prior, yet achieves measurably higher accuracy in the real world.

\subsection{Hybrid Simulator}\label{hybrid}

After the two steps above, we obtain a collision model capable of predicting the post-collision state in the real world from a given pre-collision state. We integrate this learned collision model into a physics simulator, replacing the default collision computation, to create a hybrid simulator. The simulator switches to the neural network to predict the post-collision state whenever a collision between two meshes is detected. This plug-and-play design allows seamless integration of the collision model into existing simulators with minimal modification.

\begin{figure*}
    \centering
    \begin{subfigure}{0.6\linewidth}
        \centering
        \includegraphics[width=0.98\linewidth]{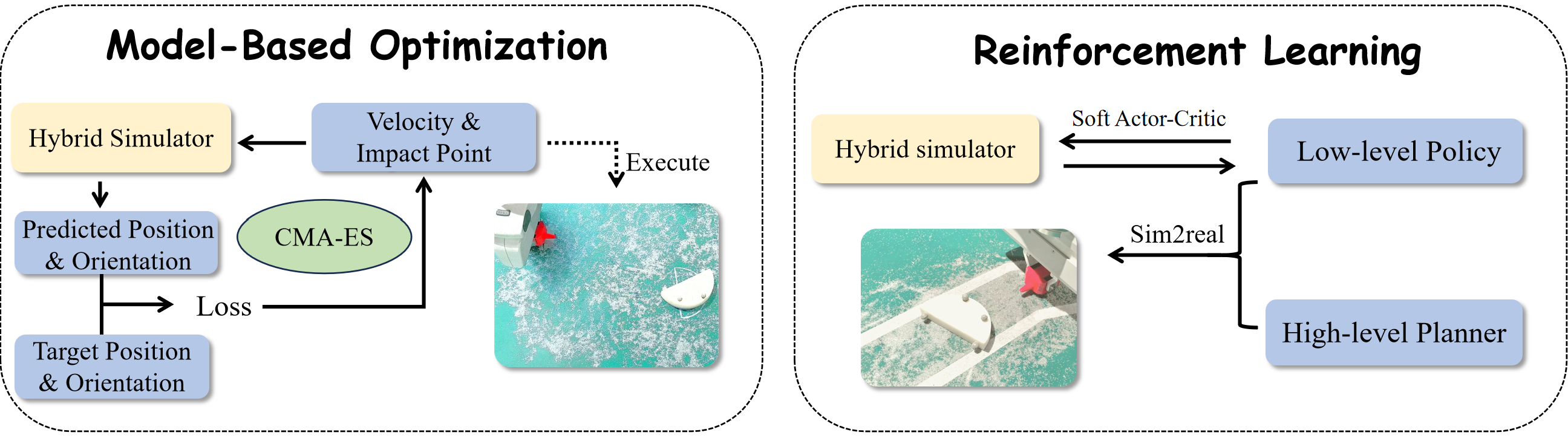}
        \caption{}
        \label{fig_downstream_task}
    \end{subfigure}
    \begin{subfigure}{0.38\linewidth}
        \centering
        \includegraphics[width=\linewidth]{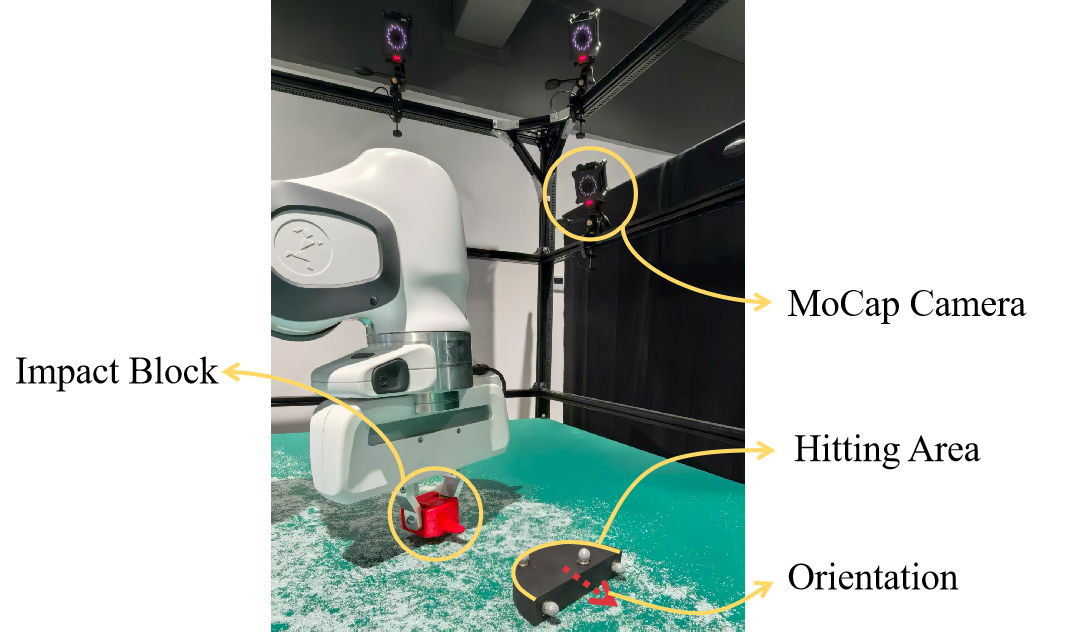}
        \caption{}
        \label{fig_system_overview}
    \end{subfigure}
    \caption{(a) The proposed hybrid simulator is applied to two downstream tasks: model-based optimization and reinforcement learning. (b) System setup, where a Franka arm manipulates an impact block to strike the semi-cylinder block.}
    \label{fig:Hybrid simulator and downstream task}
\end{figure*}

As illustrated in Fig. \ref{fig_downstream_task}, this high-fidelity hybrid simulator can be directly employed in the two main paradigms: model-based optimization and reinforecement learning (RL). First, the simulator's accuracy allows gradient-free optimization algorithms like Cross-Entropy Method(CEM)\cite{de2005tutorial} and Covariance Matrix Adaptation Evolution Strategy(CMA-ES)\cite{hansen2016cma} to discover effective open-loop control inputs that successfully transfer to the physical system for single-shot manipulation tasks. This is analogous to calculating the precise force and angle for a golf or billiard shot to achieve a desired outcome. Second, the hybrid simulator provides a physically plausible training environment that significantly improves sim-to-real policy transfer for RL. As our experiments confirm, utilizing such an accurate simulator in SAC training directly leads to policies that transfer more reliably to a real-world executions.


\section{EXPERIMENTS}

In this section, we present a series of experiments to demonstrate the effectiveness of the proposed CLASH framework. We first evaluate the prediction accuracy of the collision model trained in CLASH. Then, we construct a hybrid simulator as described in Section \ref{hybrid} and apply it to two representative downstream tasks:
\begin{itemize}
    \item finding the best control signal to hit a block given the target position and orientation,
    \item training a policy with reinforcement learning to repeatedly hit a block until it reaches the target.
\end{itemize}

These two tasks are chosen to cover both planning-based control and learning-based control paradigms. The first task highlights how the hybrid simulator can directly support model-based search for optimal actions in a single-shot hitting scenario, while the second task examines whether the improved simulator facilitates the training of robust RL policies in sequential decision-making settings.

For the simulation experiments, we adopt MuJoCo as the physics simulator since it provides the most accurate collision simulation. For real-world experiments, we use a 7-DOF Franka robotic arm to grab the impact block and hit the target block. The robot is controlled using the native libfranka C++ library in Cartesian velocity control mode. The data is recorded by a motion capture (MoCap) system with 16 cameras. The overall experimental setup is illustrated in Fig. \ref{fig_system_overview}.

\subsection{Learning Collision Model}\label{partA}
\begin{figure}
    \centering
    \includegraphics[width=0.9\linewidth]{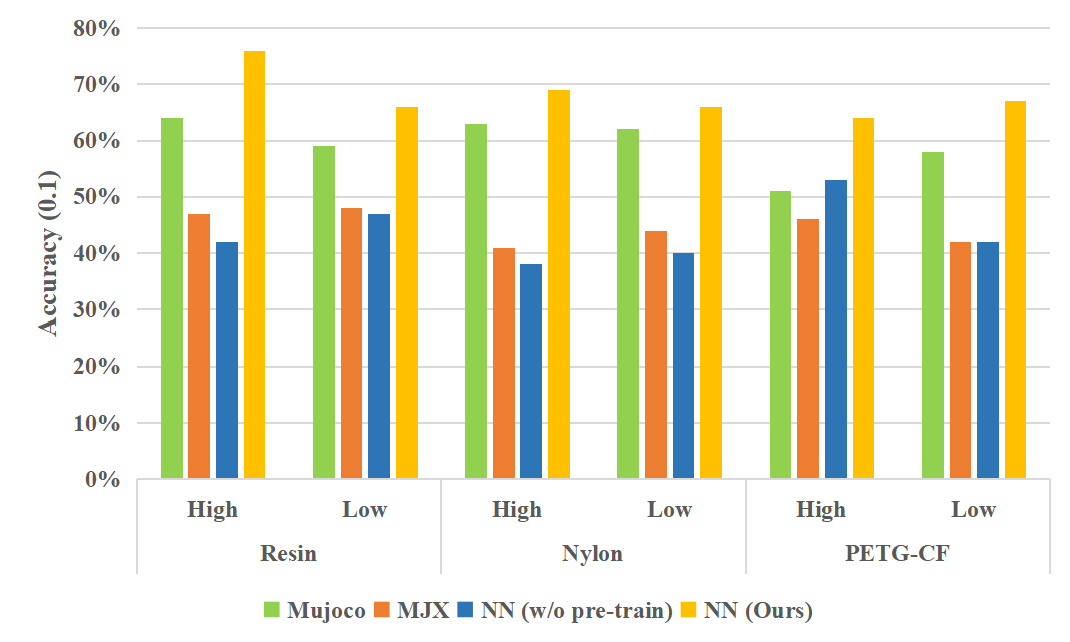}
    \caption{Collision model accuracy ($\alpha=0.1$) in different settings. Our fine-tuned strategy effectively avoids overfitting on scarce real-world data and achieves higher accuracy than MuJoCo, whose performance is limited by unmodeled effects.}
    \label{fig_collision_compare}
\end{figure}
To build a hybrid simulator, we first need to train a collision model following CLASH framework. We 3D-print a customized impact block and grab it with the Franka robotic arm to strike various stationary objects. For the target object, we primarily choose a semi-cylinder, since its anisotropic geometry makes the collision outcome highly sensitive to the impact point. 

\subsubsection{Data and Training Details}
In simulation, 100{,}000 collision pairs are generated in MuJoCo by sampling pre-collision states and system parameters (ranges in Table~\ref{table_1}). Real-world data are collected across $3$ materials $\times$ $2$ surface conditions (resin/nylon/PETG-CF; high-/low-friction sand), yielding 110 collisions per setting from varied impact angles; the first 10 samples form $D_{\text{train}}$ for system identification and adaptation, and the remaining 100 form $D_{\text{test}}$ for evaluation. The surrogate model is a three-layer MLP (width 128, ReLU). Training consists of 1{,}000 steps of simulation pre-training, followed by a two-stage real-data adaptation: optimizing contact parameters $P$ for 1{,}000 steps with frozen network weights, then fine-tuning the network for 10 steps. All runs use AdamW~\cite{loshchilov2017decoupled} with learning rate $10^{-3}$ and weight decay $10^{-2}$.
\begin{table}
\centering
\caption{Simulation Settings}
\label{table_1}
\begin{tabularx}{\linewidth}{c X c}
\toprule
Symbol & Definition & Range\\
\midrule
$v_{pre}$ & Velocity before collision & [$0.2,0.8$]\\
$\theta$ & Deflection angle of the impact point relative to the center of the semicircle & [$\frac{\pi}{12}, \frac{11\pi}{12}$]\\
$\theta_1$ & Deflection angle of the velocity relative to the impact point& [$ -\frac{\pi}{12},\frac{\pi}{12}$] \\
$\mu_1$& Friction coefficient between impact block and semi-cylinder & [$0.05,1$]\\
$\mu_2$& Friction coefficient between ground and semi-cylinder & [$0.02,1$]\\
$e_1$ & Damping ratio between impact block and semi-cylinder & [$0.1,3$] \\
$e_1$ & Damping ratio between ground and semi-cylinder & [$0.1,3$] \\
\bottomrule
\end{tabularx}
\end{table}

\subsubsection{Methods and Metrics} 
We compare our collision model against three baselines. The first baseline uses MuJoCo, which is non-differentiable. We estimate system parameters via grid search on real-world data and then simulate the collision process in MuJoCo. The second baseline employs the differentiable simulator MJX\cite{mjx_docs}. The third baseline directly fits a neural network on 10 real-world collision samples without pre-training on simulation data. For evaluation, we define prediction accuracy based on a relative error threshold. Specifically, let $D_{\alpha}$ denotes the subset of the test set $D_{\text{test}}$ in which the relative error between the predicted post-collision state and the ground-truth state is smaller than $\alpha$. The prediction accuracy is then defined as:
\begin{equation}
    \text{Accuracy} (\alpha) = \frac{|D_{\alpha}|}{|D_{\text{test}}|}
\end{equation}

\subsubsection{Results}
As shown in Fig.~\ref{fig_collision_compare}, our model consistently outperforms the three baselines across all six object–environment interaction settings. Although MJX supports efficient gradient-based parameter estimation, its applicability is restricted to simple meshes (typically fewer than 20 vertices) due to stability constraints in gradient computation. Simplifying the semi-cylinder to satisfy this requirement significantly alters its geometry, resulting in large discrepancies between simulated and real-world collisions. In contrast, our surrogate model, implemented as a three-layer MLP, avoids gradient instability and can identify collision-related parameters for complex meshes. The third baseline that directly trains on the small real-world dataset suffers from severe overfitting and poor generalization. By pre-training on large-scale simulation data and fine-tuning with limited real interactions, our model achieves substantially higher prediction accuracy and better captures the underlying collision dynamics.

\subsection{Model-Based Optimization}
\begin{table}
    \centering
    \caption{Collision Model Accuracy($\alpha=0.1$) For Three Shape Blocks.}
    \label{tab:three_shape_acc}
    \begin{tabular}{cccc}
    \toprule
       Method  & Semi-Cylinder & Square & Triangle \\
       \midrule
       MuJoCo  & 0.64& 0.50& 0.49\\
       NN(w/o Pre-train) &0.42 & 0.42&0.36 \\
       NN(ours) &\bf{0.76} &\bf{0.55} &\bf{0.54}\\
    \bottomrule
    \end{tabular}
\end{table}

\begin{table*}[htpb]
    \centering
    \caption{Error in Model-Based Optimization}
    \label{table_2}
    \resizebox{\linewidth}{!}{
    \begin{tabular}{ccccccc}
    \toprule 
    \multirow{2}{*}{} & \multicolumn{2}{c}{Semi-Cylinder} & \multicolumn{2}{c}{Square} & \multicolumn{2}{c}{Triangle} \\
    \cmidrule(lr){2-7} Simulator & Pos/m & Ori/rad & Pos/m & Ori/rad & Pos/m & Ori/rad  \\
    \midrule
    MuJoCo & 0.0245($\pm$0.0176) & 0.7882($\pm$0.6071) & 0.0222($\pm$0.0124) & 1.2696($\pm$0.6926) & 0.0251($\pm$0.0154) & \bf{0.6673($\pm$0.4036)} \\
    Hybrid Simulator (w/o Pre-train) & 0.0363($\pm$0.0156) & 0.5751($\pm$0.5074) & 0.0243($\pm$0.0140) & 1.9751($\pm$0.8621) & 0.0303($\pm$0.0187) & 0.7027($\pm$0.5069) \\
    Hybrid Simulator (Ours) & \bf{0.0159($\pm$0.0046)} & \bf{0.4152($\pm$0.3014)} & \bf{0.0163($\pm$0.0080)} & \bf{1.2479($\pm$1.0630)} & \bf{0.0185($\pm$0.0092)} & 0.7319($\pm$0.5518) \\
    \bottomrule
    \end{tabular}}
\end{table*}

After obtaining an accurate collision model, we integrate it into the physics engine to construct the hybrid simulator. The hybrid simulator employs the learned model to simulate collision outcomes, while relying on MuJoCo for other dynamics. 
\subsubsection{Experiment Setup}
In this task, we randomly select target positions and orientations, and use the simulator to search for the optimal impact point and velocity. The resulting parameters are then executed on the Franka arm in real, where we measure the distance and orientation errors between the actual and desired outcomes. We evaluate performance on three different object shapes (semi-cylinder, square, and triangle), as illustrated in Fig.~\ref{three_shape}.

\begin{table}
    \centering
    \caption{Time Cost in CMA-ES}
    \label{table_3}
    \resizebox{\linewidth}{!}{
    \begin{tabular}{cccc}
    \toprule
    Time(s) & Semi-cylinder & Square & Triangle\\
    \midrule
       MuJoCo  & 27.42($\pm$ 3.24) & 28.58($\pm$ 3.71) & 25.20($\pm$ 2.89)\\
       Hybrid Simulator (Ours) & 15.81($\pm$ 2.75) & 14.74($\pm$ 1.83)& 13.21($\pm$ 3.52)\\
    \bottomrule
    \end{tabular}}
\end{table}
\subsubsection{Methods and Metrics}


 Due to gradient instability in long-horizon rollouts and collisions with complex meshes, MJX is excluded from the baselines. We compare three simulators: MuJoCo, a hybrid simulator whose collision model is trained on real data only, and our CLASH-trained hybrid simulator. For each sampled target pose, CMA-ES~\cite{hansen2016cma} searches the impact point and velocity to minimize
\begin{equation}
    \mathcal{L}_{\text{search}}=\|\text{Pos}_{\text{pred}}-\text{Pos}_{\text{target}}\|^2 + \tau \|\text{Ori}_{\text{pred}}-\text{Ori}_{\text{target}}\|^2,
\end{equation}
with $\tau=0.01$. The optimized impact parameters are converted to Cartesian velocity commands on the Franka arm. For each shape, $5$ target poses are evaluated with $5$ trials each, and mean$\pm$std position and orientation errors are reported.

\subsubsection{Results}

\begin{figure}
    \centering
    \includegraphics[width = 0.9\linewidth]{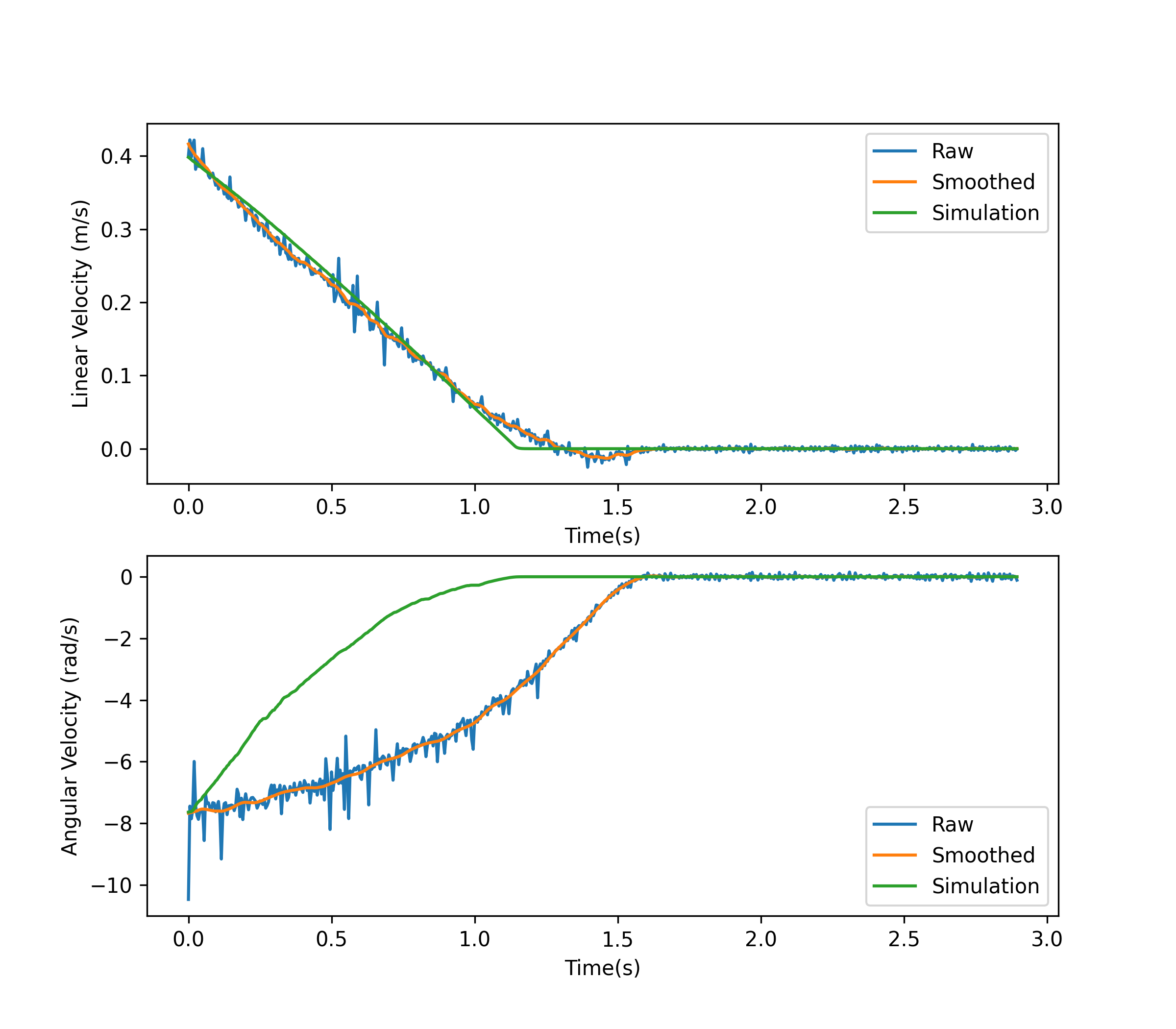}
    \caption{Linear and angular velocities of the square block in simulation and real world. MuJoCo captures the linear velocity decay reasonably well, but fails to reproduce the angular velocity decay accurately. Raw: raw measurements from the real world. Smoothed: filtered real-world data. Simulation: MuJoCo-simulated data.}
    \label{figure_compare}
\end{figure}

To build the hybrid simulators, collision models are trained for each block shape and integrated into MuJoCo. In real-world executions, the resulting hybrid simulators reduce the positioning error by 35.10\% (semi-cylinder), 26.58\% (square), and 26.29\% (triangle), as summarized in Table~\ref{table_2}. In contrast, orientation errors for the square and triangle remain large, which is likely due to inaccurate torsional-friction effects for these geometries (Fig.~\ref{figure_compare}); improving the friction model is left for future work. Table~\ref{table_3} further reports the CMA-ES wall-clock time, where replacing collision resolution with the learned model substantially accelerates collision-heavy optimization while maintaining improved accuracy.

\subsection{RL With Hybrid Simulator}

\begin{figure*}[htbp]
    \centering
    \begin{subfigure}{0.3\linewidth}
        \centering
        \includegraphics[width=\linewidth]{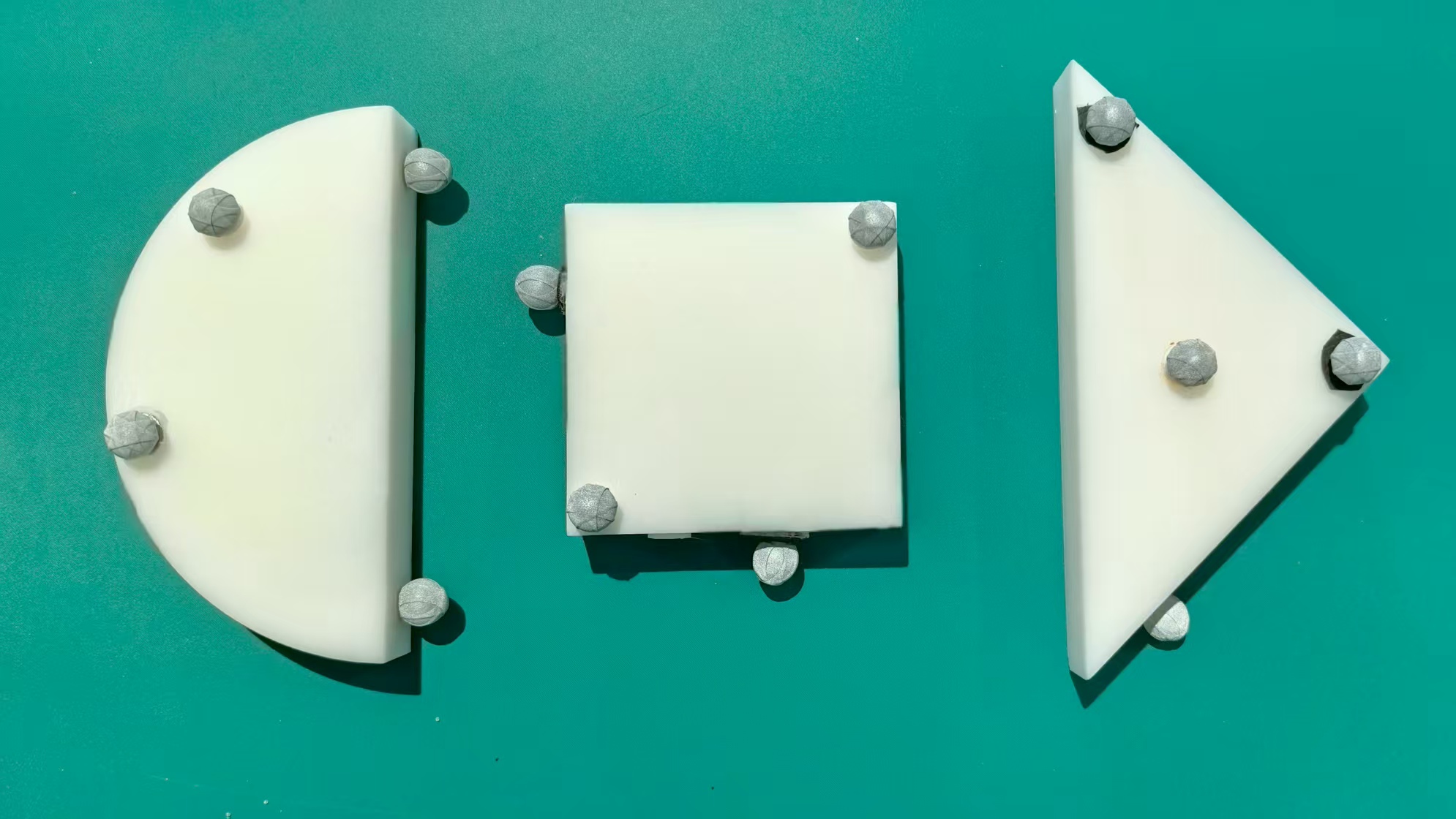}
        \caption{}
        \label{three_shape}
    \end{subfigure}
    \hfill
    \begin{subfigure}{0.3\linewidth}
        \centering
        \includegraphics[width=\linewidth]{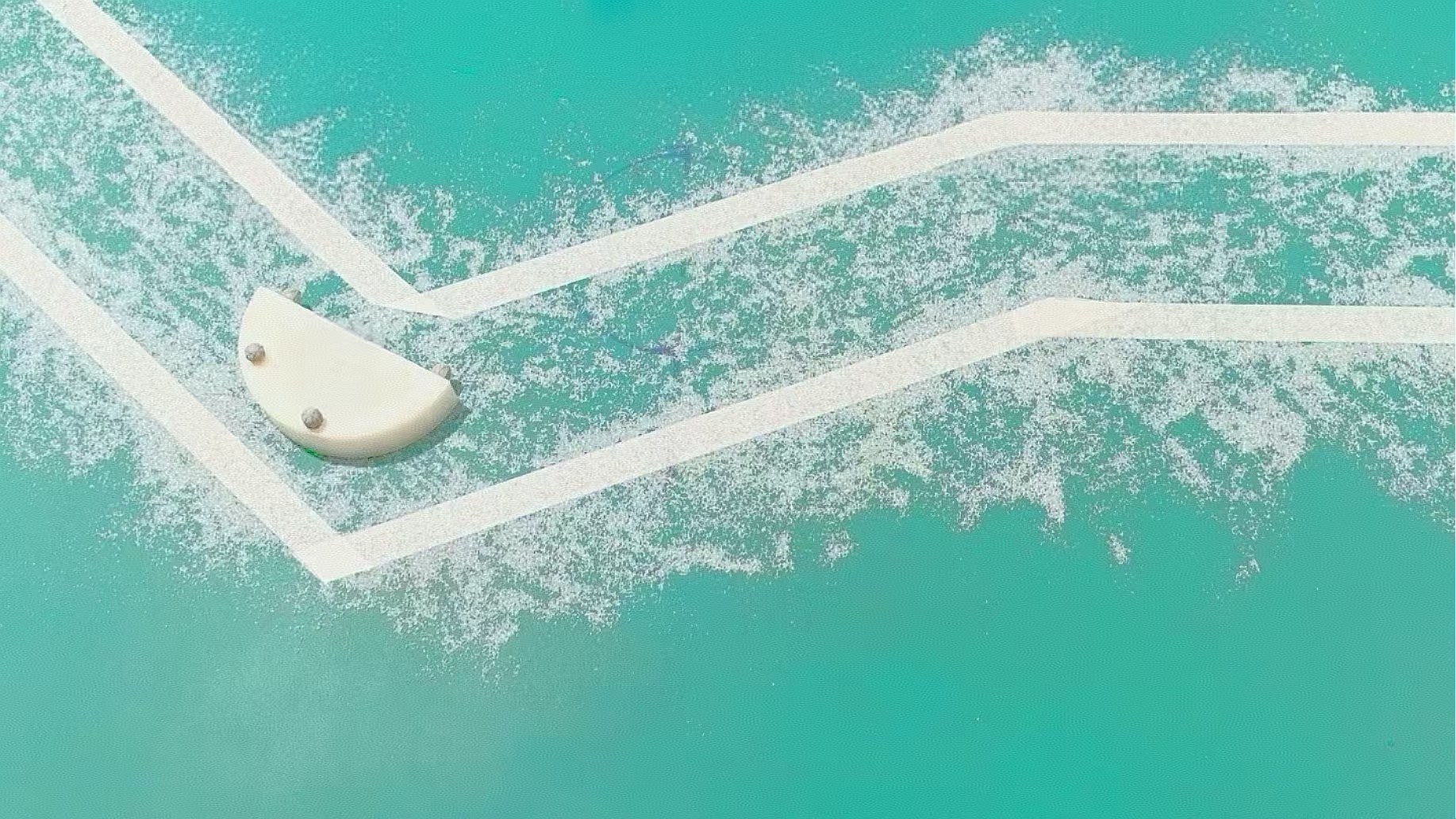}
        \caption{}
        \label{rl_task}
    \end{subfigure}
    \hfill
    \begin{subfigure}{0.3\linewidth}
        \centering
        \includegraphics[width=\linewidth]{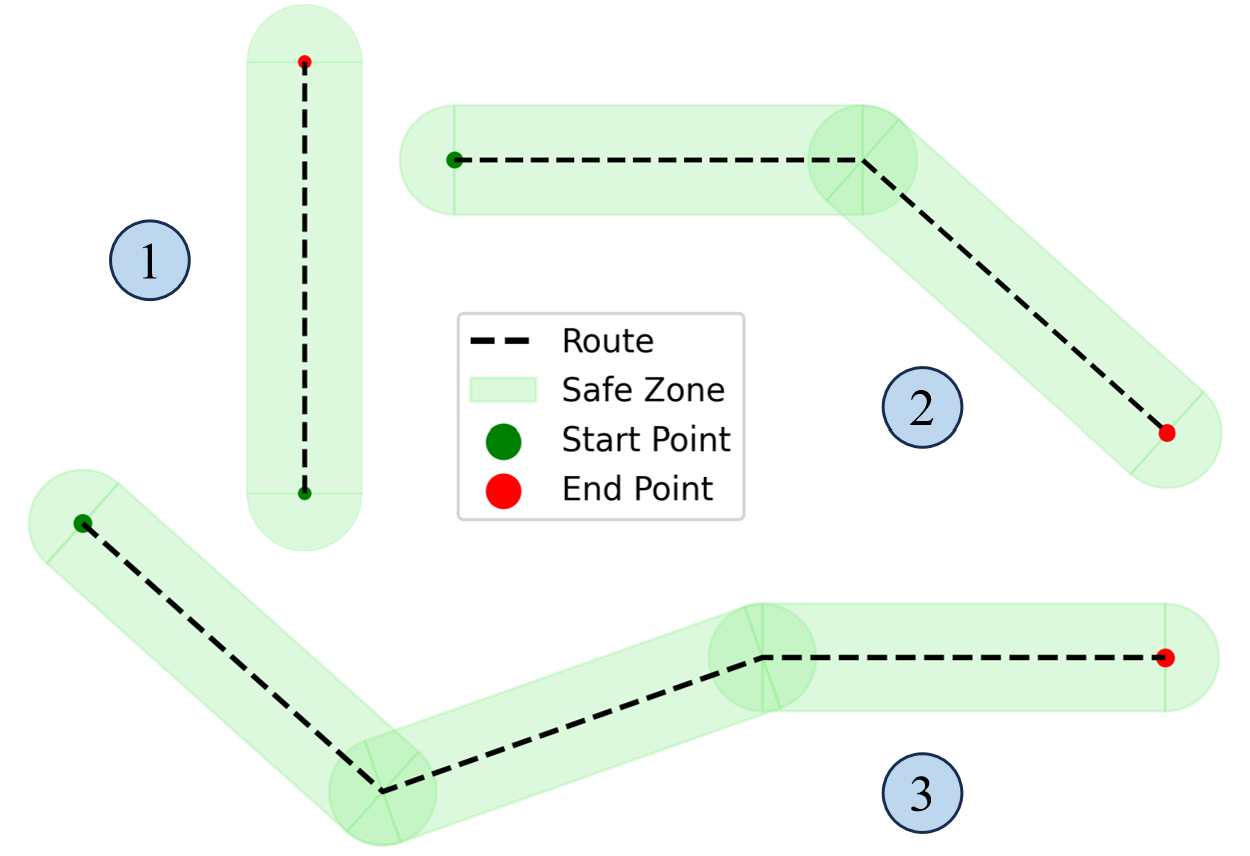} 
        \caption{}
        \label{rl_path}
    \end{subfigure}
    \caption{(a) The three block shapes used in the model-based optimization experiments: semi-cylinder, square, and triangle (from left to right).
    (b) The designated workspace, bounded by the two white lines.
    (c) The three evaluation routes used in the RL experiments. The difficulty increases from Route 1 to Route 3. For Route 1 and 2, the initial orientation of the semi-cylinder is randomized, while for Route 3 it is fixed.}
    \label{fig:three_subfigs}
\end{figure*}

\subsubsection{Experiment Setup}
We further evaluate the hybrid simulator in a reinforcement learning (RL) task. The objective is to push a semi-cylinder from a starting point to an end point without leaving the designated area (Fig. \ref{rl_task}). This task poses two main challenges. First, it requires accurate modeling of both collision and sliding, as even a single strike can cause the object to leave the safe zone, despite appearing stable in simulation. Second, only impacts on the arc side of the semi-cylinder are permitted (Fig. \ref{fig_system_overview}). If the arc is misaligned with the target direction, the strike will fail to move the object toward its destination. Consequently, orientation must be explicitly considered in the reward design. We evaluate the trained policies on routes of three different shapes (Fig. \ref{rl_path}).

\begin{table}
\centering
\caption{Reward function design for policy learning. The total reward is $R = R_1+R_2+R_3+R_4$.}
\label{tab:reward_design}
\resizebox{\linewidth}{!}{
\begin{tabular}{cll}
\toprule
Term & Formula & Description \\
\midrule
$R_1$ & $\exp(-\Delta x^2)$ & Dense position reward \\
$R_2$ & $\frac{1+\cos(\Delta \theta)}{2}$ & Orientation reward at sub-goal \\
$R_3$ & $(T - t)$ & Strike efficiency reward at sub-goal \\
$R_4$ & $r_0 $ & Penalty for leaving safe zone \\
\bottomrule
\end{tabular}}
\end{table}
\subsubsection{Method and Metrics}

We adopt a hierarchical formulation for training efficiency and composability: short straight segments are easier to learn, and arbitrarily long or diverse paths can be formed by concatenating segments. So we only train a low-level collision policy to finish one straight segment at a time. A high-level planner updates the sub-goal (position and orientation) at each segment boundary. The route is partitioned into straight segments of 0.3 $\times$ 0.08 m. A sub-goal is declared reached when the semi-cylinder is within 0.04 m of the target point. Upon completion, the next sub-goal position is set to the end of the upcoming segment, and its orientation aligns with the direction from the next sub-goal position to the one after it. This design reduces credit-assignment difficulty on long horizons while enabling straightforward path composition.


The sub-goal policy is trained with SAC in two environments (MuJoCo vs. the hybrid simulator) for 200{,}000 steps using Stable-Baselines3~\cite{stable-baselines3} (default hyperparameters). Observations include the semi-cylinder position and the sub-goal relative pose (position and orientation), while actions specify the impact point and striking velocity. The reward is $R=R_1+R_2+R_3+R_4$ (Table~\ref{tab:reward_design}): $R_1$ is a dense distance-to-goal term based on $\Delta x$, $R_2$ and $R_3$ provide sparse rewards upon reaching the sub-goal for orientation alignment and strike efficiency, and $R_4$ penalizes leaving the safe zone. Here $\Delta\theta$ is the orientation deviation, $t$ is the number of strikes used, and $T$ is the maximum allowed strikes (set to $T{=}5$); the safety penalty uses $r_0{=}{-}10$.

\begin{table}
    \centering
    \caption{Results in Sequential Semi-Cylinder Pushing Task}
    \label{table_rl}
    \begin{tabular}{cccccc}
    \toprule
      \multirow{2}{*}{} & Route 1 & \multicolumn{2}{c}{Route 2} & \multicolumn{2}{c}{Route 3}\\
    \cmidrule(lr){2-6} Simulator & SR & SR& SGCR & SR& SGCR\\
    \midrule
      MuJoCo &  4/10 & 2/10 & 8/20 & 1/10 & 13/30\\
      Hybrid Simulator (ours) &  \bf{8/10} & \bf{5/10} & \bf{14/20} & \bf{5/10} &  \bf{21/30}\\
    \bottomrule
    \end{tabular}
\end{table}

 Each path is tested ten times with the same low level policy. We report both the success rate (SR), defined as the fraction of trials in which the full path is completed, and the sub-goal completion rate (SGCR), defined as the ratio of successfully completed segments to the total number of segments. 
 
 \subsubsection{Results}
 The results are summarized in Table \ref{table_rl}. The policies trained with the hybrid simulator achieve consistently higher SR and SGCR compared to those trained with MuJoCo alone. This demonstrates that the hybrid simulator provides more accurate collision predictions, leading to policies that transfer more reliably to the real world. These findings confirm the importance of accurate collision modeling in narrowing the sim-to-real gap for RL tasks.

\section{CONCLUSIONS AND DISCUSSIONS}

In summary, CLASH provides a data-efficient framework for integrating simulation and limited real-world data to improve the modeling of contact-rich dynamics. By distilling simulator priors into a differentiable, parameter-conditioned surrogate model and adapting it with a small number of real interactions, CLASH creates a hybrid simulator that yields more accurate post-impact predictions while reducing the wall-clock cost of collision-heavy simulation and planning. More broadly, CLASH illustrates a practical pathway to enhance existing physics engines by selectively augmenting inaccurate submodules with data-driven components.

Building on this foundation, several directions remain open. First, the CLASH pipeline can be extended to other simulator submodules with known modeling gaps—such as torsional friction, rotational dissipation or compliant contact effects—by designing appropriate parameterizations and targeted data-collection protocols. Second, while CLASH distills useful priors from a base simulator, further reducing dependence on the underlying engine’s fidelity (e.g., via improved parameterizations or richer supervision signals) warrants additional study. Finally, while the current surrogate model is instantiated per geometry, incorporating scalable geometric representations (e.g., point clouds or SDFs) and shared pretraining across shapes may improve generalization and amortize per-object training, reducing the need for retraining from scratch. Overall, these directions point to a modular and reusable hybridization recipe for plug-in simulator upgrades, enabling targeted corrections where modeling errors matter most.

\bibliographystyle{IEEEtran}
\bibliography{root}




\end{document}